\documentclass[letterpaper]{article}
\usepackage{times}
\usepackage{helvet}
\usepackage{courier}
\usepackage{amsmath,amsfonts,amssymb,amsthm}
\newtheorem{definition}{Definition}
\newtheorem{proposition}{Proposition}
\newtheorem{theorem}{Theorem}
\newtheorem{remark}{Remark}

\usepackage{amsmath,amsfonts,amssymb,amsthm}
\usepackage[final]{graphics,graphicx}

\usepackage{algorithm}
\usepackage{algpseudocode}

\errorcontextlines\maxdimen

\makeatletter
    \newcommand*{\algrule}[1][\algorithmicindent]{\makebox[#1][l]{\hspace*{.5em}\thealgruleextra\vrule height \thealgruleheight depth \thealgruledepth}}%
\newcommand*{\thealgruleextra}{}
\newcommand*{\thealgruleheight}{.75\baselineskip}
\newcommand*{\thealgruledepth}{.25\baselineskip}

\newcount\ALG@printindent@tempcnta
\def\ALG@printindent{%
    \ifnum \theALG@nested>0
        \ifx\ALG@text\ALG@x@notext
        \else
            \unskip
            \addvspace{-1pt}
            \ALG@printindent@tempcnta=1
            \loop
                \algrule[\csname ALG@ind@\the\ALG@printindent@tempcnta\endcsname]%
                \advance \ALG@printindent@tempcnta 1
            \ifnum \ALG@printindent@tempcnta<\numexpr\theALG@nested+1\relax
            \repeat
        \fi
    \fi
    }%
\usepackage{etoolbox}
\patchcmd{\ALG@doentity}{\noindent\hskip\ALG@tlm}{\ALG@printindent}{}{\errmessage{failed to patch}}
\makeatother

\newbox\statebox
\newcommand{\myState}[1]{%
    \setbox\statebox=\vbox{#1}%
    \edef\thealgruleheight{\dimexpr \the\ht\statebox+1pt\relax}%
    \edef\thealgruledepth{\dimexpr \the\dp\statebox+1pt\relax}%
    \ifdim\thealgruleheight<.75\baselineskip
        \def\thealgruleheight{\dimexpr .75\baselineskip+1pt\relax}%
    \fi
    \ifdim\thealgruledepth<.25\baselineskip
        \def\thealgruledepth{\dimexpr .25\baselineskip+1pt\relax}%
    \fi
    \State #1%
    \def\thealgruleheight{\dimexpr .75\baselineskip+1pt\relax}%
    \def\thealgruledepth{\dimexpr .25\baselineskip+1pt\relax}%
}

\newcommand{\ie}{i.e.}

\newcommand{\virg}[1]{``#1''}
\newcommand{\card}[1]{|#1|}

\newcommand{\tuple}[1]{\ensuremath{\langle #1 \rangle}}
\newcommand{\set}[1]{\ensuremath{\{#1\}}}
\newcommand{\powset}[1]{\ensuremath{2^{#1}}}

\newcommand{\argument}[1]{\ensuremath{\textrm{\textbf{#1}}}}
\newcommand{\arga}{\argument{a}}
\newcommand{\argb}{\argument{b}}
\newcommand{\argc}{\argument{c}}

\newcommand{\AFname}{\ensuremath{AF}}
\newcommand{\setargs}{\ensuremath{\mathcal{A}}}
\newcommand{\setattacks}{\ensuremath{\mathcal{R}}}
\newcommand{\AF}[1]{\tuple{\setargs_{#1}, \setattacks_{#1}}}
\newcommand{\anAF}{\AF{}}
\newcommand{\anAFsymbol}{\ensuremath{\Gamma}}
\newcommand{\attacks}[2]{\ensuremath{#1 \rightarrow #2}}

\newcommand{\dungconffree}{conflict--free}
\newcommand{\dungacceptable}{acceptable}

\newcommand{\dungadmissible}{admissible}
\newcommand{\dungcomplete}{complete}
\newcommand{\dunggrounded}{grounded}
\newcommand{\dungpreferred}{preferred}

\newcommand{\aset}{\ensuremath{T}}

\newcommand{\textlabel}[1]{\ensuremath{\mathrm{\mathtt{#1}}}}
\newcommand{\textin}{\ensuremath{\textlabel{in}}}
\newcommand{\textout}{\ensuremath{\textlabel{out}}}
\newcommand{\textundec}{\ensuremath{\textlabel{undec}}}

\newcommand{\Labfun}{\ensuremath{\mathcal{L}ab}}

\newcommand{\setoflabellings}[1]{{\mathfrak{L}(#1)}}                
\newcommand{\setoflabellingsset}[1]{{\mathfrak{L}}_{#1}}          
\newcommand{\setgenlab}[2]{{\mathcal{L}}_{#1}(#2)}			 

\newcommand{\gensem}{\sigma}
\newcommand{\setgenext}[2]{\mathcal{E}_{#1}(#2)}
\newcommand{\CO}{\mathcal{CO}}
\newcommand{\GR}{\mathcal{GR}}
\newcommand{\PR}{\mathcal{PR}}

\newcommand{\attackers}[1]{#1^{-}}
\newcommand{\attacked}[1]{#1^{+}}             

\newcommand{\scc}[1]{{\mbox{SCC}}_{{#1}}}                    
\newcommand{\setbasgenn}[3]{\mathcal{BF}_{#3}(#1,#2)} 
\newcommand{\simbsetbasgenn}[1]{\mathcal{BF}_{#1}}    
\newcommand{\setextgenn}[2]{\mathcal{GF}(#1,#2)}       
\newcommand{\pundfromset}[3]{U_{#1}(#2,#3)}   

\newcommand{\restargfram}[2]{{#1}{\downarrow_{#2}}}
\newcommand{\nameinpref}{\mbox{P-PREF}}                       
\newcommand{\namepref}{\mbox{R-PREF}}                          
\newcommand{\nameprefparall}{\mbox{P-SCC-REC}}                          
\newcommand{\namebasepref}{\mbox{B-{PR}}}           
\newcommand{\namegr}{\mbox{GROUNDED}}

\newcommand{\nameboundcondII}{\mbox{L-COND}}

\newcommand{\namescctreealg}{\mbox{SCCS-LIST}}
\newcommand{\comblab}{\otimes}
\newcommand{\namegreedy}{\mbox{GREEDY}}

\newcommand{\ExtToLab}{\mathtt{Ext2Lab}}

\usepackage{authblk}

\usepackage{enumitem}
\setlist{nolistsep}

\begin{document}
%

\title{Exploiting Parallelism for Hard Problems in Abstract Argumentation: Technical Report}

\author[1]{Federico Cerutti\thanks{f.cerutti@abdn.ac.uk}}
\author[2]{Ilias Tachmazidis\thanks{ilias.tachmazidis@hud.ac.uk}}
\author[2]{Mauro Vallati\thanks{m.vallati@hud.ac.uk}}
\author[2]{Sotirios Batsakis\thanks{S.Batsakis@hud.ac.uk}}
\author[3]{Massimiliano Giacomin\thanks{massimiliano.giacomin@unibs.it}}
\author[2]{Grigoris Antoniou\thanks{g.antoniou@hud.ac.uk}}
\affil[1]{School of Natural and Computing Science\\King's College\\University of Aberdeen\\AB24 3UE Aberdeen, UK}
\affil[2]{School of Computing and Engineering\\University of Huddersfield\\HD1 3DH Huddersfield, UK}
\affil[3]{Department of Information Engineering\\University of Brescia\\via Branze, 38\\
25123, Brescia, Italy}

\renewcommand\Authands{ and }

\maketitle

\begin{abstract}
\begin{quote}
Abstract argumentation framework (\AFname) is a unifying
framework able to encompass a variety of nonmonotonic
reasoning approaches, logic programming and
computational argumentation. Yet, efficient approaches
for most of the decision and enumeration problems
associated to \AFname s are missing, thus potentially limiting
the efficacy of argumentation-based approaches in real
domains. In this paper, we present an algorithm for enumerating
the preferred extensions of abstract argumentation
frameworks which exploits parallel computation.
To this purpose, the SCC-recursive semantics definition
schema is adopted, where extensions are defined
at the level of specific sub-frameworks. The algorithm
shows significant performance improvements
in large frameworks, in terms of number of solutions
found and speedup.
\end{quote}
\end{abstract}

\section{Introduction}

Dung's theory of abstract argumentation \cite{dung1995} is a unifying framework
able to encompass a large variety of specific formalisms in the areas of
nonmonotonic reasoning, logic programming and computational argumentation.
It is based on the notion of argumentation framework (\AFname),
consisting of a set of arguments and an \emph{attack} relation between them.
Different \emph{argumentation semantics} introduce in a declarative way the criteria to determine
which arguments emerge as `justified' from the conflict,
by identifying a number of \emph{extensions}, \ie{} sets of arguments that can \virg{survive the conflict together}.
In \cite{dung1995} four \virg{traditional} semantics were introduced, namely \emph{complete}, \emph{grounded}, \emph{stable}, and \emph{preferred} semantics. 
For an introduction on alternative semantics, see \cite{KER2011}.

The main computational problems in abstract argumentation
include \emph{decision} and \emph{construction} problems,
and turn out to be computationally intractable for most of argumentation semantics \cite{dw:2009}.
In this paper we focus on the \emph{extension enumeration} problem,
\ie{} constructing \emph{all} extensions prescribed for a given \AFname: its solution  provides complete information concerning the justification status of arguments
and subsumes the solutions to the other problems.

In this paper we propose the first parallel approach for enumerating preferred extensions --- a problem which lies at the second level of the polynomial hierarchy, thus justifying the quest for efficient solutions --- which exploits the SCC-recursive schema \cite{AIJ05}, a semantics definition schema where extensions are defined at the level of the sub-frameworks identified by the strongly connected components. A similar approach has been recently discussed in \cite{KR} and compared with the state-of-the-art approach \cite{tafa}, but it does not exploit neither parallel nor dynamic programming techniques. 

As large-scale argumentation is vastly unexplored, there is no further work directly related to our approach. The closest work is in the context of Assumption-Based Argumentation (ABA) Frameworks 
 \cite{Bondarenko1997}, an abstract framework for default reasoning which can be instantiated with different deductive systems (e.g. logic programming, autoepistemic logic, default logic). \cite{Craven2012} describes a parallel implementation for credulous acceptance under the acceptablity semantics for 
some specific instances of ABAs 
in the medical domain. \cite{Craven2012}  considers competitive parallel executions: multiple versions --- equivalent w.r.t. their outcome --- of a  sequential process are created and then started in parallel. Once one version finds a solution to the problem, the others are killed. 

Our work can be seen as part of a broader recent push towards large-scale reasoning which, among others, concerns simple semantic web reasoning \cite{g1}, fuzzy ontologies \cite{g2} and logic programming \cite{g3}. Indeed, the fast-growing field of \emph{argument mining} from content in the Web \cite{Grosse2012,Cabrio2013} highlights 
 the lack of large-scale reasoning approaches in formal argumentation, and thus increases the importance of our research. 



The paper is organised as follows. In the first section we recall some necessary background on Dung's \AFname, the SCC-recursive schema and the existing algorithmic approach exploiting it. In the subsequent section we present our approach for exploiting the SCC-recursive schema in a parallel fashion, and we discuss the theoretical remarks granting the correctness of the approach. An exhaustive experimental analysis is then presented in the forthcoming section. The last section concludes the paper and discusses future work.

\section{Background}

\subsection{Dung's Argumentation Framework}

An argumentation framework \cite{dung1995} consists of a set of arguments\footnote{In this paper we consider only \emph{finite} sets of arguments: see \cite{Baroni2013} for a discussion on infinite sets of arguments.} and a binary attack relation between them.

\begin{definition}
An \emph{argumentation framework} (\AFname) is a pair $\anAFsymbol = \anAF$
where $\setargs$ is a set of arguments and $\setattacks \subseteq \setargs \times \setargs$.
We say that \argb{} \emph{attacks} \arga{} iff $\tuple{\argb,\arga} \in \setattacks$, also denoted as $\attacks{\argb}{\arga}$.
The set of attackers of an argument $\arga$ will be denoted as
$\attackers{\arga} \triangleq \set{\argb : \attacks{\argb}{\arga}}$,
the set of arguments attacked by $\arga$ will be denoted as
$\attacked{\arga} \triangleq \set{\argb : \attacks{\arga}{\argb}}$.
We also extend these notations to sets of arguments, \ie{} given $E \subseteq \setargs$,
$\attackers{E} \triangleq \set{\argb \mid \exists \arga \in E, \attacks{\argb}{\arga}}$
and
$\attacked{E} \triangleq \set{\argb \mid \exists \arga \in E, \attacks{\arga}{\argb}}$.
\end{definition}

An argument $\arga$ without attackers, \ie{} such that $\attackers{\arga} = \emptyset$,
is said \emph{initial}. Moreover, each argumentation framework has an associated directed graph where the vertices are the arguments, and the edges are the attacks.

The basic properties of \dungconffree ness, acceptability, and admissibility of a set of arguments
are fundamental for the definition of argumentation semantics.

\begin{definition}
\label{def:recall}
Given an $\AFname$ $\anAFsymbol = \anAF$:
\begin{itemize}
  \item a set $\aset \subseteq \setargs$ is a \emph{\dungconffree} set of $\anAFsymbol$
        if $\nexists~ \arga, \argb \in \aset$ s.t. $\attacks{\arga}{\argb}$;
  \item an argument $\arga \in \setargs$ is \emph{\dungacceptable} with respect to a set $\aset \subseteq \setargs$ of  $\anAFsymbol$
        if $\forall \argb \in \setargs$ s.t. $\attacks{\argb}{\arga}$,
        $\exists~ \argc \in \aset$ s.t. $\attacks{\argc}{\argb}$;
  \item a set $\aset \subseteq \setargs$ is an \emph{\dungadmissible{} set} of $\anAFsymbol$
        if $\aset$ is a \dungconffree{} set of $\anAFsymbol$ and every element of $\aset$
        is \dungacceptable{} with respect to $\aset$ of $\anAFsymbol$.
\end{itemize}
\end{definition}

An argumentation semantics $\gensem$ prescribes for any \AFname{} $\anAFsymbol$ a set of \emph{extensions}, denoted as $\setgenext{\gensem}{\anAFsymbol}$, namely a set of sets of arguments satisfying the conditions dictated by $\gensem$.
Here we recall the definitions of complete (denoted as $\CO$), grounded (denoted as $\GR$) and preferred (denoted as $\PR$) semantics. 

\begin{definition}
\label{def_sem_recall}
Given an $\AFname$ $\anAFsymbol = \anAF$:
\begin{itemize}
    \item a set $\aset \subseteq \setargs$ is a \emph{\dungcomplete{} extension} of $\anAFsymbol$,
             \ie{} $\aset \in \setgenext{\CO}{\anAFsymbol}$,
             iff $\aset$ is \dungadmissible{}
             and $\forall \arga \in \setargs$ s.t. $\arga$ is \dungacceptable{} w.r.t. $\aset$,  $\arga \in \aset$;
     \item a set $\aset \subseteq \setargs$ is the \emph{\dunggrounded{} extension} of $\anAFsymbol$,
               \ie{} $\aset \in \setgenext{\GR}{\anAFsymbol}$,
               iff  $\aset$ is the minimal (w.r.t. set inclusion) \dungcomplete{} extension of $\anAFsymbol$. Its existence and uniqueness have been proved in \cite{dungetal2006};
    \item a set $\aset \subseteq \setargs$ is a \emph{\dungpreferred{} extension} of $\anAFsymbol$,
             \ie{} $\aset \in \setgenext{\PR}{\anAFsymbol}$,
              iff $\aset$ is a maximal (w.r.t. set inclusion) \dungcomplete{} extension of $\anAFsymbol$.
\end{itemize}
\end{definition}

Each extension $\aset$ implicitly defines a three-valued \emph{labelling} of arguments:
an argument $\arga$ is labelled \textin{} iff $\arga \in \aset$;
\textout{} iff $\exists~ \argb \in \aset$ s.t. $\attacks{\argb}{\arga}$;
\textundec{} otherwise.
Argumentation semantics can be equivalently defined
in terms of labellings rather than of extensions \cite{Caminada2006,KER2011}.

\begin{definition} \label{def-gen-labelling}
  Given a set of arguments $\aset$,
   a \emph{labelling} of $\aset$ is a total function $\Labfun: \aset \longrightarrow \set{\textin, \textout, \textundec}$.
  The set of all \emph{labellings} of $\aset$ is denoted as $\setoflabellingsset{\aset}$.
  Given an \AFname{} $\anAFsymbol = \anAF$,
  a \emph{labelling} of $\anAFsymbol$ is a labelling of $\setargs$.
  The set of all \emph{labellings} of $\anAFsymbol$ is denoted as $\setoflabellings{\anAFsymbol}$.
\end{definition}

Complete labellings can be defined as follows.

\begin{definition}  \label{def:complete-labelling}
  Let $\anAFsymbol = \anAF$ be an argumentation framework.
  A labelling $\Labfun \in \setoflabellings{\anAFsymbol}$
  is a \emph{complete labelling} of $\anAFsymbol$ iff it satisfies the following conditions for any $\arga \in \setargs$:
  \begin{itemize}
  \item $\Labfun(\arga) = \textin \Leftrightarrow \forall \argb \in \attackers{\arga} \Labfun(\argb) = \textout$;
  \item $\Labfun(\arga) = \textout \Leftrightarrow \exists \argb \in \attackers{\arga}: \Labfun(\argb) = \textin$.
  \end{itemize}
\end{definition}


The grounded and preferred labelling can then be defined on the basis of complete labellings.
\begin{definition}
   Let $\anAFsymbol = \anAF$ be an argumentation framework.
   A labelling $\Labfun \in \setoflabellings{\anAFsymbol}$ is \emph{the grounded labelling} of $\anAFsymbol$
   if it is the complete labelling of $\anAFsymbol$ minimizing the set of arguments labelled \textin,
   and it is a \emph{preferred labelling} of $\anAFsymbol$ if it is a complete labelling of $\anAFsymbol$
   maximizing the set of arguments labelled \textin.
\end{definition}


The function $\ExtToLab$ provides the connection between extensions and labellings.

\begin{definition} \label{set-lab-corresp}
    Given an \AFname{} $\anAFsymbol = \anAF$ and a \dungconffree{} set $\aset \subseteq \setargs$,
    the corresponding labelling $\ExtToLab(\aset)$ is defined as $\ExtToLab(\aset) \equiv \Labfun$, where
    \begin{itemize}
       \item $\Labfun(\arga) = \textin \Leftrightarrow \arga \in \aset$
       \item $\Labfun(\arga) = \textout \Leftrightarrow \exists~ \argb \in \aset$ s.t. $\attacks{\argb}{\arga}$
       \item $\Labfun(\arga) = \textundec \Leftrightarrow \arga \notin \aset \wedge \nexists~ \argb \in \aset$
                  s.t. $\attacks{\argb}{\arga}$
    \end{itemize}
\end{definition}

\cite{Caminada2006} shows  that there is a bijective correspondence between extensions and labellings for complete, grounded, and preferred semantics.

\begin{proposition}
  Given an \AFname{} $\anAFsymbol = \anAF$,
  $\Labfun$ is a complete (grounded, preferred) labelling of $\anAFsymbol$ if and only
   if there is a complete (grounded, preferred) extension $\aset$ of $\anAFsymbol$ such that $\Labfun = \ExtToLab(\aset)$.
\end{proposition}

The set of complete labellings of $\anAFsymbol$ is denoted as $\setgenlab{\CO}{\anAFsymbol}$,
the set of preferred labellings as $\setgenlab{\PR}{\anAFsymbol}$,
while $\setgenlab{\GR}{\anAFsymbol}$ denotes the set including the grounded labelling.

\subsection{SCC-Recursiveness}   \label{sec:sccrev}

In \cite{Baroni2004}
an extension-based semantics definition schema has been introduced,
called \emph{SCC (strongly connected component)-recursiveness},
based on the graph-theoretical notion of SCCs \cite[Lemma 9]{Tarjan72} and on the observation that most argumentation semantics can be equivalently defined at the level of SCCs.

The following definitions introduce the SCC-recursive schema \cite{AIJ05}.
First, let us recall the definition of \emph{restriction} of an \AFname{} $\anAFsymbol$ to a set of arguments $I$,  in symbol $\restargfram{\anAFsymbol}{I}$.

\begin{definition}
    Given an argumentation framework $\anAFsymbol = \anAF$ and a set $I \subseteq \setargs$,
    the restriction of $\anAFsymbol$ to $I$ is defined as
    $\restargfram{\anAFsymbol}{I} \equiv (I, \setattacks \cap (I \times I))$.
\end{definition}

Then, Definition \ref{def:SCCscheme} introduces the function $\setextgenn{\anAFsymbol}{C}$
which recursively computes the semantics extensions on the basis of the SCCs of $\anAFsymbol$.
Let us denote as $\scc{\anAFsymbol}$ the set including the SCCs of an argumentation framework $\anAFsymbol$.

\begin{definition}   \label{def:SCCscheme}
   A given argumentation semantics $\gensem$ is SCC-recursive if for any argumentation framework $\anAFsymbol = \anAF$,
   $\setgenext{\gensem}{\anAFsymbol} = \setextgenn{\anAFsymbol}{\setargs} \subseteq 2^{\setargs}$.
   For any $\anAFsymbol = \anAF$ and for any set $C \subseteq \setargs$,
   $E \in \setextgenn{\anAFsymbol}{C}$ if and only if
   \begin{itemize}
       \item $E \in \setbasgenn{\anAFsymbol}{C}{\gensem}$ \mbox{if } $\card{\scc{\anAFsymbol}} = 1$
       \item $ \forall S \in \scc{\anAFsymbol} ~ (E \cap S) \in
                   \setextgenn{\restargfram{\anAFsymbol}{S \setminus \attacked{(E \setminus S)}}}{\pundfromset{\anAFsymbol}{S}{E} \cap C}$
                   otherwise
   \end{itemize}
   where
   \begin{itemize}
       \item $\setbasgenn{\anAFsymbol}{C}{\gensem}$ is a function, called \emph{base function},
       that, given an argumentation framework $\anAFsymbol = \anAF$ such that $\card{\scc{\anAFsymbol}} = 1$
       and a set $C \subseteq \setargs$, gives a subset of $2^\setargs$
       \item $\pundfromset{\anAFsymbol}{S}{E} = \set{\arga \in S \setminus \attacked{(E \setminus S)} \mid
                         \forall \argb \in (\attackers{\arga} \setminus S),~\argb \in \attacked{E} }$
   \end{itemize}
\end{definition}

The schema is based on the notions of
extension of an \AFname{} \emph{in a set of arguments}.

\begin{definition} \label{def:extinc}
  Given an  \AFname{} $\anAFsymbol = \anAF$ and a set $C \subseteq \setargs$,
  a set $E \subseteq \setargs$ is:
  an \emph{admissible set of $\anAFsymbol$ in $C$}  if and only if
  $E$ is an admissible set of $\anAFsymbol$ and $E \subseteq C$;
  a \emph{complete extension of $\anAFsymbol$ in $C$} if and only if
  $E$ is an admissible set of $\anAFsymbol$ in $C$, and
  every argument $\alpha \in C$ which is acceptable with respect to $E$ belongs to $E$;
  the \emph{grounded extension of $\anAFsymbol$ in $C$} if and only if it is the least (w.r.t. set inclusion) 
  complete extension of $\anAFsymbol$ in $C$;
  a \emph{preferred extension of $\anAFsymbol$ in $C$} if and only if it is a maximal (w.r.t. set inclusion)
  complete extension of $\anAFsymbol$ in $C$.
\end{definition}

The existence and uniqueness of the grounded extension in $C$, as well as the existence of at least a preferred extension in $C$, have been proved in \cite{AIJ05}.
Moreover, 
\cite{AIJ05} proves that $\setextgenn{\anAFsymbol}{C}$, as defined in  Def. \ref{def:SCCscheme},
returns the $\gensem$-extensions in $C$
(with $\gensem \in \set{\CO, \GR, \PR}$), provided that $\setbasgenn{\anAFsymbol}{C}{\gensem}$
returns the complete, grounded, and preferred extensions in $C$, respectively.

\cite{KR} introduces the notions of  complete, grounded and preferred labellings of $\anAFsymbol$ in $C$, 
\ie{} the labelling-based counterparts of  the corresponding notions of Definition \ref{def:extinc}, 
and describes a preliminary algorithm --- \namepref{} ---  exploiting the SCC-recursive schema. \namepref{}  implements $\mathcal{GF}$ (Def. \ref{def:SCCscheme}), where the chosen base function $\simbsetbasgenn{\PR}$ is computed by a refinement of the algorithm in \cite{tafa} which exploits a SAT solver as a NP-oracle to determine the preferred labellings.
\namepref{} exploits the SCC-recursive schema by constructing a sequence of strongly connected components of
$\anAFsymbol$ in a topological order. 
Preferred labellings are incrementally constructed along the SCCs,
by computing the preferred labellings of each SCC and merging them with those identified in the previous SCCs.
In the following, we take advantage of two algorithms mentioned in \cite{KR}, namely \namegr\ \cite[Alg. 3]{KR} and \namebasepref\ \cite[Alg. 4]{KR}: their usage is described in the following section.

\section{Exploiting Parallel Computation}

In this section we present our approach exploiting parallel computation in the context of the SCC-recursive schema. First of all, we need to identify when it is possible to parallelise the process aimed at verifying that given $\anAFsymbol = \AF{}$,  
$\forall E \subseteq \setargs, \forall C \subseteq \setargs, E \in \setextgenn{\anAFsymbol}{C}$.

\subsection{Theoretical Remarks}

Two elements guaranteeing independence and thus the possibility to parallelise the process can be identified. First of all, each preferred extension can be computed independently from the others.

\begin{remark}
\label{remark:parall-extensions}
Given an $\anAFsymbol = \AF{}$, $\forall E \in \setgenext{\gensem}{\anAFsymbol}, \forall C \subseteq \setargs$, proving that $E \in \setextgenn{\anAFsymbol}{C}$ does not require any knowledge about $\overline{E} \in \setgenext{\gensem}{\anAFsymbol}\setminus\set{E}$.
\end{remark}

A second, rather more articulated, condition of independence requires to identify two sets of SCCs, $\overline{S} = \set{S_1, \ldots, S_n} \subseteq \scc{\anAFsymbol}$ and $P_{\overline{S}} = \set{P_1, \ldots, P_m} \subseteq \scc{\anAFsymbol}$ such that (1) each SCC in $\overline{S}$ does not attack the others in $\overline{S}$; and (2) each SCC in $\overline{S}$ is attacked only by SCCs in $P_{\overline{S}}$.

\begin{remark}
\label{remark:parall-tree}
Given an $\anAFsymbol = \AF{}$, $\forall E \subseteq \setargs, \forall C \subseteq \setargs$, if there exist $\overline{S} = \set{S_1, \ldots, S_n} \subseteq \scc{\anAFsymbol}$ such that $\forall S_i, S_j, S_i^+\cap S_j = \emptyset$, 
and there exists $P_{\overline{S}} \subseteq \scc{\anAFsymbol}$ such that 
$\forall S_i \in \overline{S}, ({S_i}^- \setminus S_i) \subseteq \bigcup_{P \in P_{\overline{S}}} P$,
then $\forall S \in \overline{S}$ proving that
$(E \cap S) \in
                   \setextgenn{\restargfram{\anAFsymbol}{S \setminus \attacked{(E \setminus S)}}}{\pundfromset{\anAFsymbol}{S}{E} \cap C}$
 can be determined in function of $P_{\overline{S}}$ and does not require any knowledge about $S' \in \overline{S}\setminus\set{S}$.
\end{remark}

\subsection{The $\nameprefparall$ Algorithm}

In this section we introduce a meta-algorithm --- \nameprefparall\ (Alg. \ref{alg:parallel}) --- which exploits the SCC-recursive schema using parallel computation  and a pro-active \emph{greedy} approach which \emph{memoizes} some notable cases.

First of all, the function  $\nameinpref$ (Algorithm \ref{alg:base-p}) receives as input an argumentation framework
$\anAFsymbol = \anAF$ and returns the set of preferred labellings of $\anAFsymbol$.
This is simply achieved by invoking (at line $3$) $\nameprefparall(\anAFsymbol,\setargs)$,
where the function $\nameprefparall$ ($\mathcal{GF}$ in Def. \ref{def:SCCscheme}) receives as input an argumentation framework $\anAFsymbol = \anAF$
and a set $C \subseteq \setargs$, and computes the set $\setgenlab{\PR}{\anAFsymbol, C}$,
\ie{} the set of preferred labellings of $\anAFsymbol$ in $C$.

\nameprefparall\ first pre-processes (at line $3$) --- via the function \namegr\  \cite[Alg. 3]{KR} --- $\anAFsymbol$ by computing the grounded labelling in $C$: $\Labfun$ contains the restriction of the grounded labelling to those arguments which are either \textin\ or \textout; $U$ is the set of arguments that are labelled \textundec\ in the grounded labelling.

At line $4$ \nameprefparall\ initialises to $\set{\Labfun}$ the variable $E_p$, which stores the set of labellings that are incrementally constructed. At line $5$ \nameprefparall\  restricts $\anAFsymbol$ to $\restargfram{\anAFsymbol}{U}$.
Then, at line $6$, \nameprefparall\ exploits Remark \ref{remark:parall-tree} by building a list $L := (L^1, \ldots, L^n)$ 
of sets of SCCs --- \cite[p. 617]{Cormen2009} with some modifications --- such that 
$\forall L^i \in L, L^i = \set{S_j^i \in \scc{\anAFsymbol} ~\mid~ (S_j^i)^- \setminus S_j^i \in \bigcup_{z \in \set{1, \ldots, i-1} }\bigcup_{S \in L^{z}} S ~\mbox{and}~(S_j^i)^+ \setminus S_j^i \in \bigcup_{z \in \set{i+1, \ldots, n} }\bigcup_{S \in L^{z}} S}$.

At line $7$,  the \namegreedy\ function  (Alg. \ref{alg:greedy}) is called; it receives as input the list of SCCs $L$ and the set of arguments $C$, and returns a set $M$ of pairs $(S_i, B_i)$ where $S_i \in \scc{\anAFsymbol}$, 
and $B_i = \setgenlab{\PR}{\restargfram{\anAFsymbol}{S_i}, S_i \cap C}$. 
$B_i$ --- computed by the function \namebasepref\ \cite[Alg. 4]{KR} --- is the set of preferred labellings 
for $S_i$ when no argument in $S_i$ is attacked by $\textin$ or $\textundec$ arguments in previous, w.r.t. the $L$ list, SCCs.

\begin{algorithm}[t]
  \caption{Computing \dungpreferred{} labellings of an \AFname}   \label{alg:base-p}
  \textbf{$\nameinpref(\anAFsymbol)$}
  \begin{algorithmic}[1]
  \myState{\textbf{Input:} $\anAFsymbol = \anAF$}
  \myState{\textbf{Output:} $E_p \in \powset{\setoflabellings{\anAFsymbol}}$}
  \myState{\textbf{return $\nameprefparall(\anAFsymbol,\setargs)$}}
  \end{algorithmic}
\end{algorithm}

\begin{algorithm}[H]
   \caption{Computing \dungpreferred{} labellings of an \AFname\ in $C$}   \label{alg:parallel}
   \textbf{$\nameprefparall(\anAFsymbol,C)$}
  \begin{algorithmic}[1]
    \myState{\textbf{Input:} $\anAFsymbol = \anAF$, $C \subseteq \setargs$}
    \myState{\textbf{Output:} $E_p \in \powset{\setoflabellings{\anAFsymbol}}$}
    \myState{$(\Labfun, U) = \namegr(\anAFsymbol, C)$}
    \myState{$E_p~:=~\set{\Labfun}$}
    \myState{$\anAFsymbol = \restargfram{\anAFsymbol}{U}$}
    \myState{$\begin{array}[t]{@{} l @{} l}
        L & :=  (L^1 := \set{S^1_1, \ldots, S^1_k}, \ldots, L^n := \set{S^n_1, \ldots, S^n_h})\\
            & ~= \namescctreealg(\anAFsymbol)
      \end{array}$}
    \myState{$M~:=~ \set{\ldots, (S_i, B_i), \ldots} = \namegreedy(L, C)$}
    \For{$l \in \set{1, \dots,n}$}
        \myState{$E_l~:=~\set{E_l^{S_1} := (), \ldots, E_l^{S_k} := ()}$} 
        \For{$S \in L^l$}  \underline{\textbf{in parallel}}
          \For{$\Labfun \in E_p$}  \underline{\textbf{in parallel}}
              \myState{$(O, I)~:=~\nameboundcondII(\anAFsymbol, S, L^l, \Labfun)$}
              \If{$I = \emptyset$}
                   \myState{$
                     \begin{array}[t]{@{} l @{} l}
                       E_l^S[\Labfun] = & \set{\set{(\arga, \textout) \mid \arga \in O} ~\cup \\
                                       & \set{(\arga, \textundec) \mid \arga \in S \setminus O}}
                     \end{array}$}
              \Else
                  \If{$I = S$}
                           \myState{$E_l^S[\Labfun] = B \mbox{ where }(S, B) \in M$}
                  \Else
                       \If{$O=\emptyset$}
                             \myState{$E_l^S[\Labfun] =  \namebasepref(\restargfram{\anAFsymbol}{S}, I \cap C)$}
                       \Else    
		             \myState{$E_l^S[\Labfun] \! = \! \set{\set{(\arga, \textout) \mid \arga \in O}}$}
                             \myState{$
                                     \begin{array}[t]{@{} l @{} l}
                                        E_l^S[\Labfun] & = E_l^S[\Labfun] \comblab \\
                                            & \nameprefparall(\restargfram{\anAFsymbol}{S \setminus O}, I \cap C)
                                      \end{array}$}
                       \EndIf
                  \EndIf
              \EndIf
          \EndFor
        \EndFor
        \For{$S \in L^l$} 
          \myState{$E'_p~:=~\emptyset$}
          \For{$\Labfun \in E_p$} \underline{\textbf{in parallel}}
              \myState{$E'_p = E'_p \cup (\set{\Labfun} \comblab E_l^S[\Labfun])$}
          \EndFor
          \myState{$E_p~:=~E'_p$}
        \EndFor 
    \EndFor
    \myState{\textbf{return} $E_p$}
  \end{algorithmic}
\end{algorithm}

Then, at lines $8-36$, \nameprefparall\ performs a first loop among the elements of the list $L$.
At line $9$ a rather articulated data structure, $E_l$, is initialised. 
For each $S \in L_l$, $E_l^S$ is a list of pairs $(\Labfun, E_l^{S}[\Labfun] \subseteq \setoflabellingsset{S})$: to ease of notation, hereafter we omit the pair-structure  thus referring directly to $E_l^{S}[\Labfun]$ which contains the set of preferred labellings of $S$ 
constructed on the basis of a specific labelling $\Labfun$ identified in
the previous (w.r.t. the list $L$) SCCs.

Two more loops are thus considered and their execution can be safely parallelised: the loop at lines $10-28$ exploits Remark \ref{remark:parall-tree} by considering each SCC in a given element of the list $L$; while the loop at lines $11-27$ considers a single preferred labelling, each of which is independent from the others --- cf. Remark \ref{remark:parall-extensions}.

$\nameboundcondII(\anAFsymbol, S, L^l, \Labfun)$ 
at line $12$ computes the effect of previous SCCs, and returns $(O, I)$, where:
\begin{itemize}
\item $O = \set{\arga \in S \mid \exists \argb \in \aset \cap \attackers{\arga} : \Labfun(\argb)=\textin}$ and
\item $I = \set{\arga \in S \mid \forall \ \argb \in \aset \cap \attackers{\arga} , \Labfun(\argb)=\textout}$,
\end{itemize}

\noindent
with  ${\aset \equiv \bigcup_{i = 1}^{l-1} \bigcup_{S \in L_i} S}$.
Variable $O$ is set to include arguments of $S$ that are attacked by
``outside'' $\textin$-labelled arguments according to $\Labfun$,
and variable $I$ is set to include arguments of $S$ that are only attacked by ``outside''  $\textout$-labelled arguments.
This gives rise to three cases:
\begin{enumerate}
\item each argument of $S$ is attacked by $\textin$ or $\textundec$ arguments in previous SCCs --- hence each argument of $S$ is labelled \textout\ or \textundec\ (line $14$);
\item no argument of $S$ is attacked by $\textin$ arguments in previous SCCs: this is the base case of the recursion and thus either we exploit the \emph{memoization} technique implemented with the \namegreedy\ algorithm (line $17$)
or we exploit the function \namebasepref\ (line $20$);
\item in the remaining case, arguments attacked by \textin-arguments are labelled as $\textout$ and \nameprefparall\ is recursively called on the restriction of $S$ to the unlabelled arguments (lines $22-23$).
\end{enumerate}

Finally, at lines $29-35$ the computed preferred labellings $E_l^S[\Labfun]$ are merged together ($E_1 \comblab E_2 = \set{\Labfun_1 \cup \Labfun_2 | \Labfun_1 \in E_1, \Labfun_2 \in E_2}$) with the $\Labfun$ labelling of previous SCCs. Once again, due to Remark \ref{remark:parall-extensions}, this process can be parallelised (lines $31-33$).

Then the algorithm considers the next element in the list $L$. Once the outer loop is exited,
 all strongly connected components have been processed,
thus $E_p$ is returned as the set of preferred labellings in $C$ (line $37$).

\begin{algorithm}[H]
  \caption{Greedy computation of base cases}   \label{alg:greedy}
   \textbf{$\namegreedy(L, C)$}
  \begin{algorithmic}[1]
  \myState{\textbf{Input:} $L = (L^1, \ldots, L^n := \set{S_1^n, \ldots, S_h^n}), C \subseteq \setargs$}
    \myState{\textbf{Output:} $M = \set{\ldots, (S_i, B_i), \ldots}$}
    \myState{$M := \emptyset$}
    \For{$S \in \bigcup_{i =1}^n L^i$}  \underline{\textbf{in parallel}}
        \myState{$B~:=~\namebasepref(\restargfram{\anAFsymbol}{S}, S \cap C)$}
        \myState{$M = M \cup \set{(S, B)}$}
    \EndFor
    \myState{\textbf{return} $M$}
  \end{algorithmic}
\end{algorithm}

\begin{theorem}
  Given an \AFname{} $\anAFsymbol = \anAF$ and a set $C \subseteq \setargs$, 
  Algorithm \ref{alg:parallel} returns $E_p = \setgenlab{\PR}{\anAFsymbol, C}$. 

  \begin{proof}
    This follows from \cite[Thm. 1]{KR} and Remarks \ref{remark:parall-extensions} and \ref{remark:parall-tree}.
  \end{proof}
\end{theorem}

\section{Empirical Analysis}


The solvers have been run on a cluster with computing nodes equipped with 2.4 Ghz Dual Core AMD Opteron{\texttrademark}, 8 GB of RAM and Linux operating system. As in the International Planning Competition (IPC) \cite{Jimenez2012}, a cutoff of 900 seconds was imposed to compute the preferred extensions for each \AFname. No limit was imposed on the RAM usage, but a run fails at saturation of the available memory.
Moreover, we adopted the IPC speed score, also borrowed from the planning community, which is defined as follows.
For each  \AFname, each system gets a score of $1/(1 + \log_{10}(T/T^*))$, where $T$ is its execution time and $T^*$ the best execution time
among the compared systems, or a score of 0 if it fails in that case. Runtimes below 0.01 sec get by default the maximal score of 1. In our experimental analysis, IPC score is normalised to 100. 
For each solver we recorded the overall result: success (if it finds each preferred extension), crashed, timed-out or ran out of memory.

As shown in \cite{comma14a}, most of the state-of-the-art approaches for enumerating preferred extensions hardly solve large (w.r.t. the number of arguments) frameworks. In this work, we focus on extremely large \AFname s; the largest -- as far as we know -- that have ever been used for testing solvers.

We randomly generated a set of 200 \AFname s, varying the number of SCCs between 90 and 210, the number of arguments between 2,700 and 8,400, and considering different uniformly distributed probabilities of attacks, either between arguments or between different SCCs, leading to \AFname s with a number of attacks between approximately 100 thousands and 2 millions. \AFname s were generated using {\tt AFBenchGen} \cite{comma14b}.

In the following we rely on the Wilcoxon Signed-Rank Test (WSRT) in order to identify significant subset of data \cite{Wilcoxon1945}. 

\begin{table}[!t]

\begin{center}
\begin{tabular}{ | l | c | c  c  c  c  |}
\hline

& P1 & P2 & P2G & P4 & P4G \\
 \hline
IPC score & 50.0 & 58.6 & 44.7 & {\bf 74.9} & 56.9\\
\% success & 58.8 & 68.3 & 57.3 & {\bf 74.9} & 65.8 \\
\% best & 1.0 & 0.0 & 0.0 & {\bf 74.4} & 0.0 \\
Avg runtime & 439.4 & 464.3 & 385.7 & {\bf 269.6} & 384.8\\
Speedup & -- & 1.9 & 1.3 & 2.8 & 1.7 \\
\hline

\hline

\end{tabular}

\end{center}
\caption{\label{tab:cfr} Performance achieved by using 1, 2 and 4 processors with/out exploiting the \emph{greedy} approach (+G). Results are shown in terms of normalised IPC score, percentages of success, percentages of \AFname s in which the system has been the fastest, average runtime (considering \AFname s in which at least one approach succeeded) and max speedup against P1. Values in bold indicate the best results.}
\end{table}

Table \ref{tab:cfr} shows the results of the overall comparison between R-PREF (henceforth P1) and $\nameprefparall$ 
(henceforth P2 or P4, according to the number of processors). The latter exploits either two or four processors, and has been run with and without the \emph{greedy} approach. From Table \ref{tab:cfr}, two main conclusions can be derived. First, the exploitation of \emph{greedy} approach introduces a significant overhead, due to the required pre-calculation. In our testing instances, pre-calculated knowledge is not used by algorithms and therefore, exploiting a \emph{greedy} approach has a detrimental effect on P2 and P4 performance. This behaviour is confirmed also by a comparison (not shown) between P1 with/out \emph{greedy} approach. Given this result, the \emph{greedy} approach will not be considered in the rest of this section.

The second conclusion we derive from Table \ref{tab:cfr} is that parallelisation improves significantly the performance of both runtime and the number of successfully analysed \AFname s. Since the normalised IPC score of P4 is equal to its percentage of successes, P4 is always the fastest approach on the whole testing set. Both P4 and P2, according to WRST perform significantly better than P1 ($p < 0.05$).
Using 2 (resp. 4) processors provides a maximum speed-up of 1.9 (resp. 2.8) times w.r.t. serial execution. Such results justify the use of parallel approaches in abstract argumentation.

\begin{figure}[H]
  \centering
  \includegraphics[width=\linewidth]{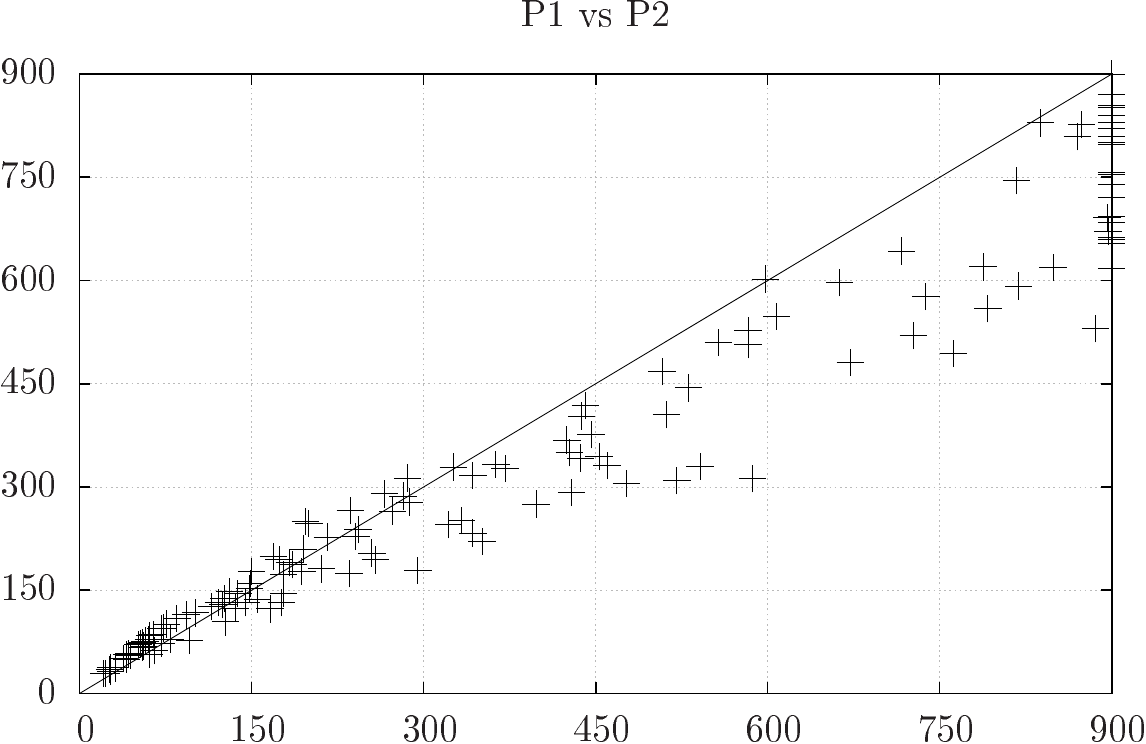}

\vspace{0.5cm}

  \includegraphics[width=\linewidth]{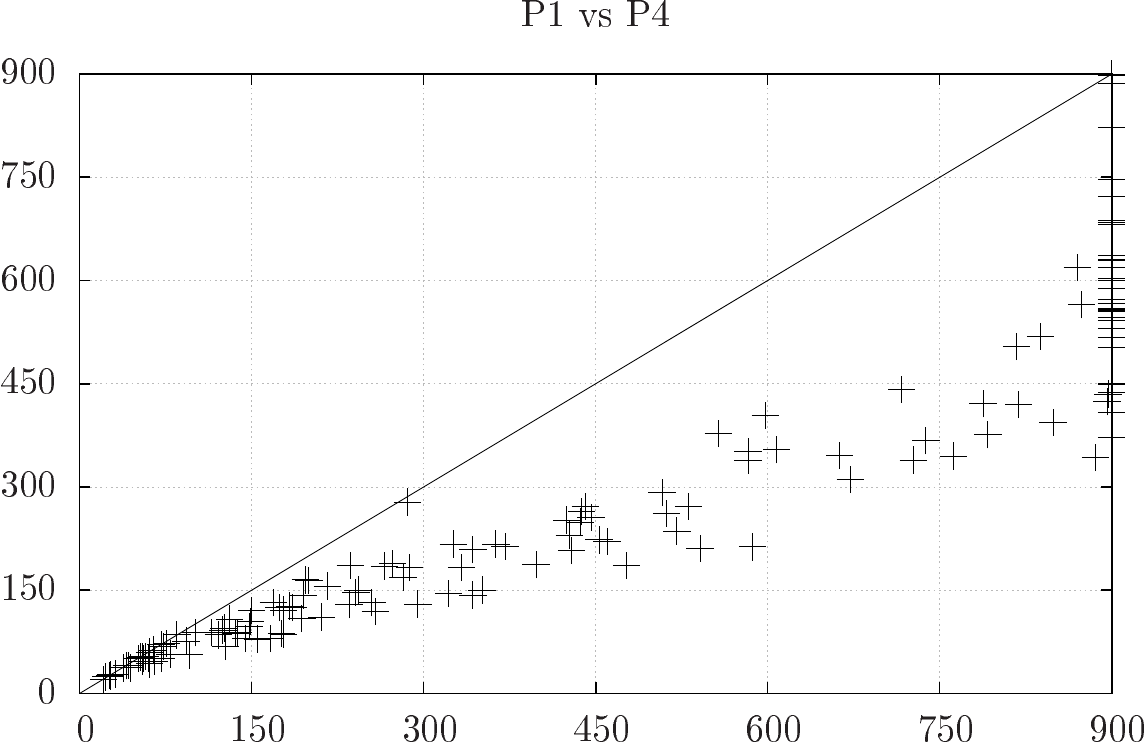}
  \caption{CPU-time of P2 (upper) or P4 (lower) w.r.t. P1 for all the considered \AFname s. The x-axis refers to CPU seconds of P1; the y-axis refers to CPU seconds of P2 (upper) or P4 (lower). CPU-time of 900 seconds indicates timeout.}
  \label{fig:scatter}
\end{figure}

\begin{figure}[H]
  \centering
  \includegraphics[width=\linewidth]{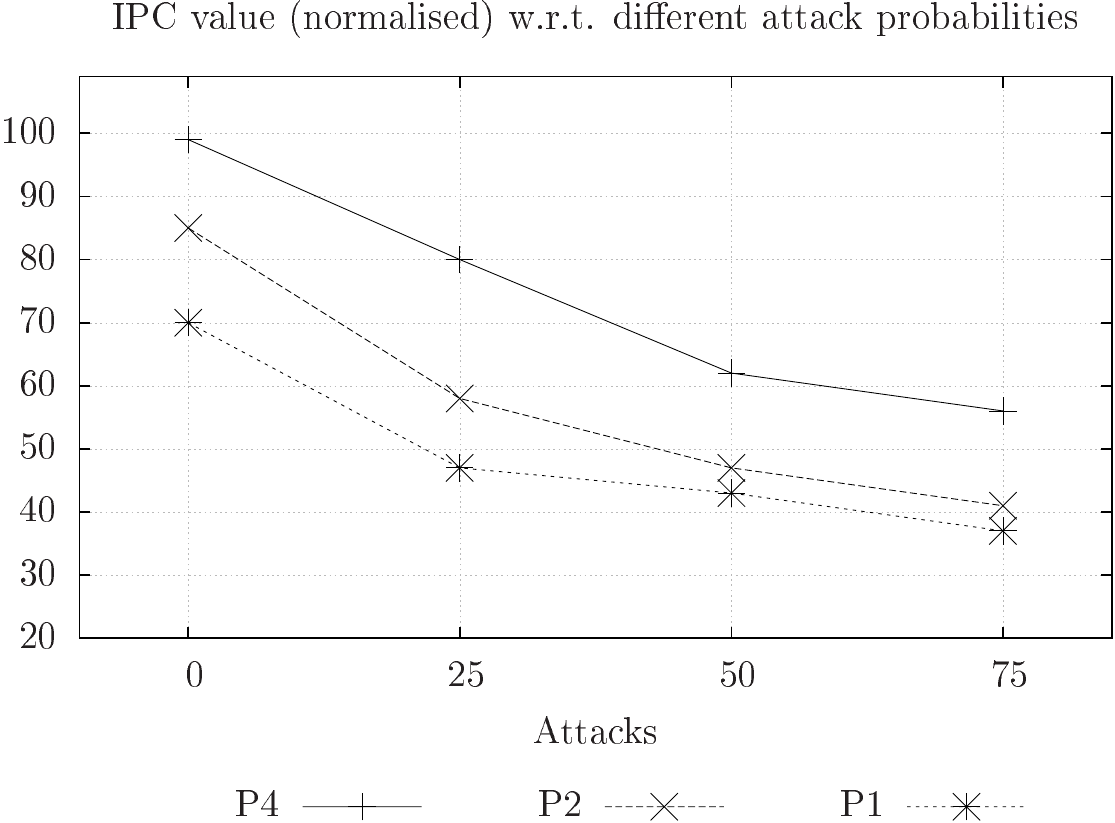}

\vspace{0.5cm}

  \includegraphics[width=\linewidth]{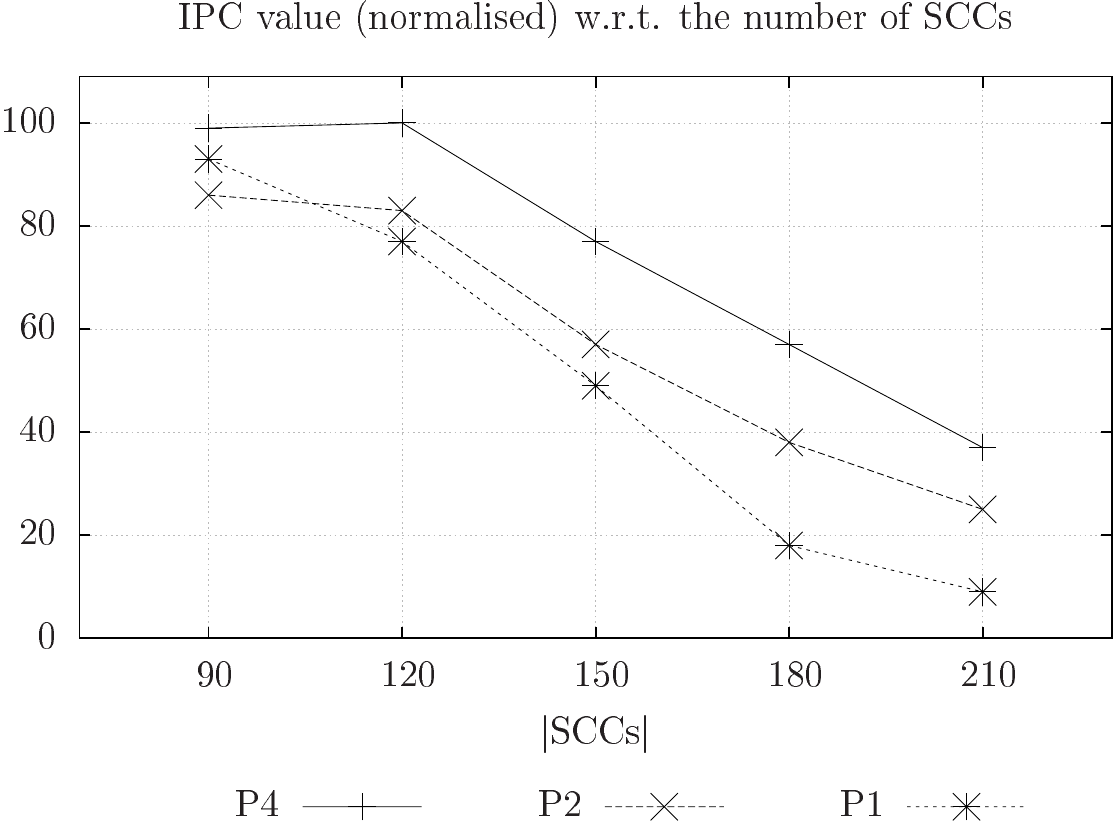}
  \caption{IPC scores of P2, P4 and P1 w.r.t. the probability of attacks between different SCCs (upper) and the number of SCCs (lower).}
  \label{fig:sccs}
\end{figure}

Figure \ref{fig:scatter} provides results in the form of scatterplots, showing the performance of P1 and, respectively, P2 and P4. Using 2 processors have a remarkable impact on runtimes, in particular on complex \AFname s, which require approximately more than 300 seconds. A larger number of \AFname s can be successfully analysed by P2: this can be derived by observing the elements on the right axis of the graph. On \emph{simple} \AFname s, the impact of using 2 processors is not so clear. On the other hand, parallelising on 4 processors guarantee to obtain lower runtimes on the whole testing set. This behaviour is probably due to the fact that the overhead introduced by parallelisation (generating threads, communication overhead, etc.) is not completely compensated by using 2 processors only, specially when a short amount of CPU time is needed for enumerating the extensions of a given \AFname.

The number of SCCs in the same set, cf. Remark \ref{remark:parall-tree}
, critically affects the performance of the proposed parallel algorithm. The larger the size of each level, the higher the degree of parallelisation that can be reached, since parallelisation is primarily based on processing simultaneously SCCs that are located on the same level. 
Figure \ref{fig:sccs} (upper) show the IPC score of parallelised and serial algorithms, with regards to the probability of attacks between SCCs. As expected, the performance gap between parallelised (P2, P4) and serial (P1) algorithms is maximum when the probability is 0 --- i.e., all the SCCs are on the same level --- and slowly decreases as the percentage increases. With a probability of 75\%, most of the levels have a single SCC, therefore parallelisation does not provide a great speedup.
It is worthy to notice that at higher attacks probability percentages, enumerating all the preferred extensions is very complex, and requires a significant amount of CPU-time. The differences of performance between P1 and P4 are always statistically significant ($p < 0.05$). It is not the case of P1 and P2, their performance are statistically indistinguishable on sets with probability of attacks of 50\% ($p = 0.39$) and 75\% ($p=0.66$).


Finally, Figure \ref{fig:sccs} shows how IPC score of considered algorithms changes with regard to the number of SCCs of the \AFname s. As a general trend, increasing the number of SCCs increases the runtime (and decreases the number of successes) for all implementations. This is expected, as larger inputs are harder to solve. On the other hand, P1 is very quick on smallest considered \AFname s; on average it is faster than P2. P1 performance rapidly decreases as the number of SCCs increases. This is also confirmed by the WRST: while P4 is always statistically better than P1, P2 performs statistically worse than P1 on \AFname s with $|$SCCs$|=90$, but it performs statistically better when $|$SCCs$| >= 120$. Generally, parallelisation provides best speedup on very large \AFname s, with lower probability of attacks among SCCs.


\section{Conclusions}

In this paper we proposed an approach for exploiting the SCC-recursive schema for computing semantics extensions in Dung's \AFname s taking advantage of parallel executions and dynamic programming.
It is worth mentioning that Alg. \ref{alg:base-p}, in conjunction with Algs. \ref{alg:parallel} and \ref{alg:greedy}, are meta-algorithms that implement the SCC-recursive schema independently from the chosen semantics. Although we chose to consider the preferred semantics in order to provide a direct comparison with recent works \cite{KR} --- in essence equivalent to P1 --- , the same algorithms can work for each semantics that is SCC-recursive \cite{AIJ05}.

Moreover, the empirical analysis shows that there is a substantial statistically significant increment of performance due to the partial parallel execution of the proposed algorithms. This results in:
\begin{enumerate}
\item an increment (approx. 50\%) of the number of \AFname s for which we can solve the preferred semantics enumeration problem before the chosen cutoff time;
\item a significant speedup of the computation of preferred extension up to 280\% just considering 4 processors.
\end{enumerate}

Future work is already envisaged in the area of additional experimentation analyses by considering different benchmarks and by discussing a variation of Alg. \ref{alg:parallel} where 
the two inner loops of Alg. \ref{alg:parallel} (lines $10-24$ and $11-23$) are swapped. 
We also plan to apply more dynamic programming techniques (e.g. \emph{memoization}) by improving the current proposal of the \emph{greedy} computation of some preferred labelling --- Alg. \ref{alg:greedy}.
In addition, we will compare our approach with \cite{Dvorak2011,Ellmauthaler2014}, which reduce the problem of enumerating preferred extensions to an ASP program which can be solved using the parallel ASP solver \texttt{clingo}. 
Finally, recent works on Input/Output behaviour characterisation of \AFname s \cite{Baroni2012,Baroni2014} can be exploited for determining conditions of independent computation and thus exploiting parallel executions on graph structures different from SCCs. 







\bibliographystyle{alpha}
\bibliography{ref}

\end{document}